%% file: ms.tex
\documentclass{article}
\PassOptionsToPackage{numbers, compress}{natbib}
\usepackage[nonatbib, final]{neurips_2022}
\usepackage[numbers]{natbib}
\usepackage{microtype}
\usepackage{float}
\usepackage[pdftex]{graphicx}
\usepackage{subfigure}
\usepackage{booktabs}
\usepackage{makecell}
\usepackage[utf8]{inputenc}
\usepackage[T1]{fontenc}    
\usepackage{hyperref} 
\usepackage{url}            
\usepackage{amsfonts}       
\usepackage{nicefrac}       
\usepackage{microtype}      
\usepackage[
  font = small,
  labelfont = bf,
  tableposition = top
]{caption}
\usepackage{amsfonts, amssymb, amsthm, amsmath}
\usepackage{wrapfig}        
\usepackage[ruled]{algorithm2e}
\usepackage{listings}       
\usepackage{bm}             
\usepackage{xcolor}
\usepackage{tikz}
\usepackage{tabu}
\usepackage{makecell}
\usepackage{multicol}
\usepackage{colortbl}
\usepackage{bbm}
\usepackage{multirow}
\usepackage[normalem]{ulem}
\usepackage{cleveref}
\usepackage{threeparttable}
\usepackage{lipsum}
\usepackage{pifont}
\usepackage[makeroom]{cancel}
\usepackage{mathtools}
\input{math_commands.tex}

\newcommand{\titlename}{Masked Autoencoding for Scalable \\ and Generalizable Decision Making}
\newcommand{\ours}{{MaskDP}}
\title{\titlename}
\author{
  Fangchen Liu\textsuperscript{1 *} \quad 
  Hao Liu\textsuperscript{1 *} \quad 
  Aditya Grover\textsuperscript{2} \quad 
  Pieter Abbeel\textsuperscript{1}
\vspace{0.1cm} \\ 
\textsuperscript{1} Berkeley AI Research, UC Berkeley \quad \textsuperscript{2} UCLA 
\vspace{0.1cm} \\
\textsuperscript{*} Equal contribution 
\vspace{0.1cm} \\ 
\texttt{\{fangchen\_liu, hao.liu\}@berkeley.edu} \\[2mm]
}
\begin{document}
\maketitle

\begin{abstract}
We are interested in learning scalable agents for reinforcement learning that can learn from large-scale, diverse sequential data similar to current large vision and language models. To this end, this paper presents masked decision prediction (MaskDP), a simple and scalable self-supervised pretraining method for reinforcement learning (RL) and behavioral cloning (BC). In our MaskDP approach, we employ a masked autoencoder (MAE) to state-action trajectories, wherein we randomly mask state and action tokens and reconstruct the missing data. By doing so, the model is required to infer masked-out states and actions and extract information about dynamics. We find that masking different proportions of the input sequence significantly helps with learning a better model that generalizes well to multiple downstream tasks. In our empirical study, we ﬁnd that a MaskDP model gains the capability of zero-shot transfer to new BC tasks, such as single and multiple goal reaching, and it can zero-shot infer skills from a few example transitions. In addition, MaskDP transfers well to offline RL and shows promising scaling behavior w.r.t. to model size. It is amenable to data-efficient finetuning, achieving competitive results with prior methods based on autoregressive pretraining\footnote{The implementation of \ours{} is available at~\url{https://github.com/FangchenLiu/MaskDP_public}}. 
\end{abstract}

\section{Introduction}
%




Self-supervised pretraining has made tremendous successes for unsupervised representation learning in natural language processing (NLP) and vision~\citep{he2021masked, devlin2018bert, bao2021beit, brown2020language}.
These methods work by predicting a removed portion of the data, which is often referred to as masked token prediction.
By varying the masking patterns and architectures, different methods have been developed for NLP and vision, $e.g.$, Transformer~\citep{vaswani2017attention}, GPT~\citep{brown2020language}, BERT~\citep{devlin2018bert}, and MAE~\citep{he2021masked}. 
These methods are simple to implement and scalable to large Internet-scale datasets and deep neural networks, leading to excellent flexibility and generalization for downstream tasks~\citep{devlin2018bert, he2021masked, alayrac2022flamingo}.  

In this work, we explore the generality of masked token prediction for generalizable and flexible reinforcement learning (RL). 
Prior work has explored sequence modeling for sequential decision making in the context of offline RL $e.g.$ decision transformer (DT)~\citep{chen2021decision} and trajectory transformer (TT)~\citep{janner2021offline}, and black-box optimization $e.g.$ transformer neural processes (TNP)~\citep{nguyen2022transformer}.
These methods are based on autoregressive next token prediction, similar to GPT~\citep{brown2020language}.
While promising, these works do not leverage diverse unlabeled data for generalization across various downstream tasks. 
In addition, DT~\citep{chen2021decision} needs reward-labeled high quality datasets, while TT~\citep{janner2021offline} requires discretizing states and actions, further limiting its applicability. 
The flexibility of applying arbitrary masks for executing various task specifications in RL significantly lags behind NLP and vision. 




We propose Masked Decision Prediction (\ours{}), a pretraining method to learn generalizable models that achieve data-efficient adaptation to various downstream tasks. 
\ours{} is a self-supervised pretraining method that can leverage unlabeled diverse data. With \ours{} pretraining, the model can generalize well to both goal reaching and offline RL, two distinctive and popular RL paradigms.

Our first key observation is that masked token prediction with random masking similar to BERT~\citep{devlin2018bert} and MAE~\citep{he2021masked} provides a general and flexible way for learning from unsupervised data. 
Unlike autoregressive action prediction used in prior works, random masking is strictly more general and requires the model to infer masked out states and actions, and thus leads to a single model that can reason about both the forward and inverse dynamics from each sample.  

Our second key observation 
is that since states and actions are highly correlated temporally, trajectories have significantly lower information density, $i.e$, it is easier to predict action or state based on nearby states and actions. 
Consequently, a high mask ratio ($95\%$) is necessary to make reconstruction task meaningful.
Unlike in MAE~\citep{he2021masked} and BERT~\citep{devlin2018bert} where the goal is learning representations, we want to directly apply \ours{} to various downstream tasks, and different mask ratios induce different pre-train and downstream gaps. For example, consider the goal-reaching task within certain time limit. Given current state, future goal and mask tokens between them, the model should be able to inpaint intermediate actions as the goal-reaching plan. The mask ratio varies from short-term plans to long-term plans. Therefore, we combine multiple different mask ratios ($e.g.$ $15\%$, $35\%$, $75\%$, and $95\%$), and mask a portion of data using a randomly sampled mask ratio. 
Our experiments show that doing so is crucial to achieving high performance. We show that self-supervised pretrained \ours{} achieves high performance in challenging multiple goals reaching setting, outperforming strong baselines in a zero-shot manner. 

We highlight our key results here: 
\begin{itemize}
    \item \textbf{Single goal reaching}: \ours{} achieves performance that exceeds or matches both training from scratch task-specific methods and other pretraining based methods. 
    \item \textbf{Sequential multiple goal reaching}: \ours{} can reach a sequence of goals effectively, even without closed-loop execution, while outperforming iterative baselines significantly. 
    \item \textbf{Offline RL}: \ours{} achieves competitive results as specialized approaches. Notably, we demonstrate that non-autoregressive architecture works well for offline RL tasks. 
\end{itemize}

\begin{figure}[!t]
    \centering
    \includegraphics[width=.99\textwidth]{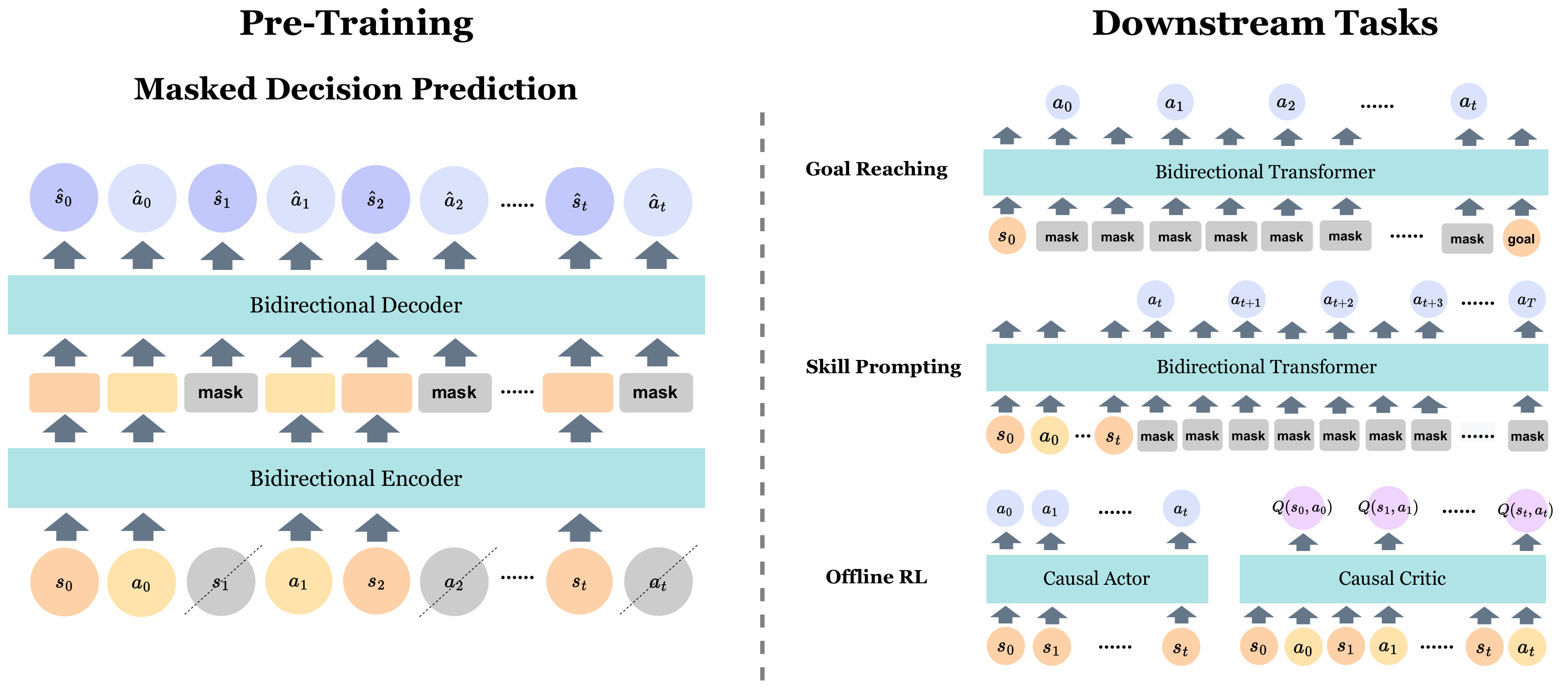}
    \caption{Illustration of MaskDP. During pretraining stage, we perform the masked token prediction task. And after pretraining, the model can be deployed to various downstream tasks using different mask patterns.
    }
    \label{fig:mdp}
\end{figure}

\section{Related work}
\paragraph{Masked modeling in language and vision.}
Large-scale language models are highly successful~\citep{devlin2018bert, brown2020language}--after pretraining on a large amount of data, these pretrained representations generalize well to various downstream tasks.
Taking inspiration from the success in NLP, Transformer~\citep{vaswani2017attention} based methods have been proposed to model images~\citep{chen2020generative, dosovitskiy2020image, bao2021beit, he2021masked}. iGPT~\citep{chen2020generative} operates on sequences of pixels and predicts unknown pixels. 
BEiT~\citep{bao2021beit} proposes to predict discrete tokens~\citep{van2017neural, ramesh2021zero}. MAE~\citep{he2021masked} proposes to randomly mask patches of the input image and reconstruct the missing pixels. 
Since we apply random mask across states and actions, our work is also related to prior work on masked prediction across multiple input modalities~\citep[see e.g.][]{wang2020vd}.


\paragraph{Sequence modeling in RL}
Sequence modeling for decision making has been studied in RL and other applications~\citep{pathak2017curiosity, schwarzer2020data, chen2021decision, janner2021offline, zheng2022online, fang2019smt}.
MVP~\citep{xiao2022masked} studies transferring pretrained visual representations to RL tasks.
MWM~\citep{seo2022masked} studies masked prediction over convolutional features, and learn a latent dynamics model.
Modeling inverse dynamics has also been studied for robot learning from demonstrations and sim-to-real transfer~\citep{christiano2016transfer, torabi2018behavioral}.
TT~\citep{janner2021offline} studies autoregressive next token prediction for model-based RL applications.
DT~\citep{chen2021decision, zheng2022online} study masking autoregressive next token prediction conditioned on return.
TNP~\cite{nguyen2022transformer} and BONET~\cite{krishnamoorthy2022generative} study autoregressive masking for sequential decision making for black-box optimization.
ICM~\citep{pathak2017curiosity} and SPR~\citep{schwarzer2020data} study predicting masked state and action in a transition tuple for exploration. 
Different from these works, \ours{} randomly masks a portion of trajectories and generalizes prior masking strategies such as inverse dynamics. In addition, \ours{} generalizes well to downstream tasks while prior work is task-specific.
Concurrent to our work, Uni[MASK]~\citep{carroll2022uni} proposes using the bidirectional transformer to predict masked states and actions and demonstrates that the resulting model performs well on a large variety of tasks. The main difference between our works is that Uni[MASK] is more interested in comparing the performance between training regimes with different masking schemes, while our work focuses on solving a range of downstream tasks with minimal task-specific designs during pretraining.

\paragraph{Unsupervised pretraining in RL}
Our work falls under the category of self-supervised pretraining in RL.
Self-supervised discovery of a set of task-agnostic behaviors by means of seeking to maximize an intrinsic reward has been explored as intrinsic motivation~\citep{baldassarre2013intrinsically}, often with the goal of encouraging exploration~\citep{csimcsek2006intrinsic, oudeyer2009intrinsic}. 
APT~\citep{liu2021behavior} studies nonparametric entropy maximization for pretraining and is extended to learning skills~\citep{liu2021aps}.
Proto-RL~\citep{yarats2021reinforcement} further improves pretraining by representation learning.
CIC~\citep{laskin2022cic} combines contrastive learning with skill discover and improves results on URL benchmarks~\citep{laskin2021urlb}.
APV~\citep{seo2022reinforcement} shows successful transfer of pretrained representation across domains. 
Many of these methods are used to pretrain agents that are later adapted to specific reinforcement learning tasks. Using offline data for pretraining agents has also been explored in prior work~\citep{schwarzer2021pretraining, stooke2021decoupling, zhan2020framework}. SGI~\citep{schwarzer2021pretraining} proposes combining self-predictive representation~\citep{schwarzer2020data} and inverse dynamics prediction. ATC~\citep{stooke2021decoupling} studies contrastive pretraining on trajectories and shows transferring the representations to downstream tasks.


\section{Method}
The key idea in \ours{} is to mask and reconstruct state-action sequences during the pretraining stage. Post pretraining, \ours{} can be zero-shot deployed or finetuned for various downstream tasks. The paradigm of the model for pretraining and finetuning is summarized in~\Figref{fig:mdp}.

\subsection{\ours{} Pretraining}
\paragraph{Random masking.} 

For sequences with low information density, a \textit{high} masking ratio is typically applied to eliminate information redundancy and make the task sufficiently difficult to avoid trivial interpolation from visible neighbor tokens. However, unlike vision and language, where the goal is to learn good representations; we also consider directly deploying this model by leveraging its inpainting capability for various downstream tasks. For example, we can give the model a goal at timestep $T$ and mask all the future inputs, the model can generate intermediate actions by inpainting the mask tokens. The mask ratio varies from goal to goal, depending on the time budget. To reduce the gap between training and deployment, we keep a set of mask ratios ($i.e.$ $15\%$, $35\%$, $50\%$, $75\%$, and $95\%$), and the data is randomly masked with a ratio sampled from this set. We find that masking multiple proportions of the input yields a meaningful self-supervisory task. 

We apply random masking on state tokens and action tokens independently. By doing so, the model is implicitly learning both forward and inverse dynamics. This also provides more flexibility as we can provide state or action-level inputs but not transition-level.

\paragraph{Architecture} 
Our encoder is a Transformer~\citep{vaswani2017attention} but applied only on visible, unmasked states and actions, similar to MAE~\citep{he2021masked}.
The states and actions are first embedded by separated linear layers, positional embeddings are then added, and lastly, the embeddings are processed by a series of self-attentional blocks. The decoder operates on the full set of encoded visible state and action tokens and mask tokens. Each mask token is a shared, learned vector that indicates the presence of a missing token to be predicted. Similar to the encoder, the masked whole sequence will pass through separated linear projections added with positional embedding prior to being passed to the decoder. Both the encoder and decoder are bidirectional.

\paragraph{Prediction target}
Our \ours{} reconstructs the input by predicting the whole action and state sequences. The last layer of the decoder consists of two MLPs to decode states and actions separately. The loss function computes the mean squared error (MSE) between the reconstructed whole sequence and original inputs. Different from other masked prediction variants~\cite{he2021masked, devlin2018bert}, we found mask loss is not useful in our setting, as our goal is to obtain an scalable decision making model but not only for representation learning.


\subsection{\ours{} Downstream Tasks}
\label{sec:tasks}
\paragraph{\ours{} for goal reaching}
We consider the problem of reaching one goal or multiple goals from a given state. The model has to generate a sequence of actions to reach goals within a certain amount of steps. \ours{} denoising pretraining objective fits the goal reaching scenario well as the model must learn to inpaint masked actions based on remaining states. 
In this task, the \ours{} encoder input is a concatenation of initial state and goals, and the decoder input is a concatenation of initial state embedding, masked token sequence, and goal embeddings. Note that the number of masked tokens determines the number of timesteps the model is expected to reach the given goals. The model then generates a state-action sequence, where we can directly execute the whole action sequence (namely "open-loop"), or only execute the first action and forward the model again with the obtained new observation (namely "closed-loop").


\paragraph{\ours{} for skill prompting}
Skill prompting requires the model to generate a trajectory conditioned a given context. For example, consider a walker agent: if we prompt it with a few state-action pairs of walking/running/standing, it should continue to generate a trajectory in the same skill pattern. 
Accordingly, we append the observed initial state-action sequences with masked tokens for the future.
The model can be rolled out once to generate the whole future sequence, or queried repeatedly to refill the masked tokens at each time step. Similar to goal-reaching task, we refer to these strategies as "open-loop" and "closed-loop" respectively.

\paragraph{\ours{} for offline RL}
In offline RL, the objective is to learn one model for maximizing the return for a task specified by a reward function. This is different from our self-supervised pretraining target, so extra finetuning is needed.  We adopt a standard actor-critic framework similar to TD3~\cite{fujimoto2018addressing} by adding a critic head and actor head, where the actor takes a state sequence as input, and the critic takes the state-action sequence as input. Both are mask-free. To match the setting in RL, we change the bidirectional attention mask in the transformer to a causal attention mask. More details about RL finetuning can be found in section~\ref{sec: rl}.




\section{Experiments}
In our experiments, we evaluate transfer learning in downstream tasks using \ours{}. 
\Secref{sec:exp_setup} introduces the environments, pretraining, and the baselines compared in experiments. 
\Secref{sec:exp_result} summarizes the results of \ours{} on goal reaching, skill prompting, and offline RL. 
Through further analysis in~\Secref{sec:exp_analysis}, we present an ablation study on various design choices of our model.

\subsection{Experiments Setup}
\label{sec:exp_setup}
\paragraph{Environments: domains vs. tasks}
We adopt the environment setup used in EXoRL~\citep{yarats2022dont}, based on DeepMind control suite~\citep{tassa2018deepmind}, where a domain describes the type of agent (e.g. Walker) but tasks are specified by rewards (e.g., Walker walk, Walker run). We use 3 domains (Walker, Cheetah and Quadruped) with 7 tasks in total. More details about the environments can be found in the Appendix.

\setlength{\leftmargini}{0.7cm}
\paragraph{Pretraining datasets}
Real-world pretraining data generally varies greatly in quality. To mimic this, we construct two different pretraining datasets to approximate different data quality scenarios. 
\begin{itemize}
    \item \textbf{Near-expert}: For every task, we train a TD3 agent~\citep{fujimoto2018addressing} for 1M steps and freeze its parameters. We rollout the policy with Gaussian random noise and collect 4M experience on each task.
    \item \textbf{Mixed}: This dataset consists of diverse data collected from various agents, including 2K near-expert trajectories for each task. Similar to ExoRL~\citep{yarats2022dont}, we collect 10M exploratory trajectories using intrinsic reward from Proto-RL~\citep{yarats2021reinforcement} for each domain. We also to use a TD3~\citep{fujimoto2018addressing} agent to maximize the sum of extrinsic reward and the Proto-RL intrinsic reward, and store its 2M experience on each task.
\end{itemize}
For more details about the above datasets and more ablations on the dataset quality, please refer to~\Secref{sec:exp_detail} and~\Secref{sec:mixv2} respectively. We perform both single-task and multi-task pretraining using the above datasets. The former leverages task-specific data while the latter utilizes data from all tasks within the same domain. 
We pretrain agents for 400K gradient steps. Specifically, for the model pretrained on the near-expert dataset, we perform zero-shot\footnote{{We directly evaluate the model on some unseen state-goal pairs in the validation dataset}} evaluation of goal reaching and skill prompting, and finetuning for offline RL; for model trained on the mixed dataset, we provide the finetuning results in~\Secref{sec:scalability} and~\Secref{sec:exp_detail}. 

\paragraph{Baselines}
\begin{itemize}
    \item GPT. We train an autoregressive model similar to GPT~\citep{brown2020language} which takes the past states and actions as input to predict the next state or action. 
    \item Goal-GPT. We specifically modify GPT to Goal-GPT to evaluate its performance on goal reaching tasks. The model takes current goal and observations as input, and predicts the action to reach this goal. The model is trained using a behaviour cloning loss as~\citep{chen2021decision}.
    \item Goal-MLP. Standard behavior cloning method that conditions on the goal. The major difference between this and Goal-GPT is here we do not use the causal Transformer architecture to make the history visible. 
\end{itemize}

By default, \ours{} uses a 3-layer encoder and 2-layer decoder, and the baselines based on GPT use 5 attention layers. \ours{} and all the above models are comparable with similar architecture design and size, and share the same training hyper-parameters. Details about the architecture and training of \ours{} and the above baselines can be found in~\Secref{sec:exp_detail}.


\subsection{Main Results}
\label{sec:exp_result}
\subsubsection{Goal Reaching} 

We consider both single and multiple goal-reaching settings. The agent is required to reach one or multiple goals from a given state, which are all sampled from the same trajectory to guarantee reachabilty within a reasonable time budget. During evaluation, the agent rolls out to reach the given goal(s) within a time budget. The evaluation dataset is also collected by the same RL agent in 3 environments with different seeds, which is unseen during pretraining. The detailed settings are:
\begin{itemize}
    \item \textbf{single-goal reaching}: For every trajectory in the validation set, we randomly sample a start state and a future state in $T \in [15, 20)$ steps as the goal. All the methods are evaluated on the same set of 300 state-goal pairs with a given budget of $T+3$. We set the agent to the start state and report the L2 distance between the goal and the closest rollout state within this budget. 
    
    \item \textbf{multi-goal reaching}: For every trajectory in the validation set, we randomly sample a start state and 5 goal states at random future timesteps from $[12, 60)$. We evaluate the same set of 100 state-goal sequences and add additional 5 timestep budgets for all the goals. Similar to single-goal reaching, We report the L2 distance between every goal and the closest rollout state before running out of its corresponding budget.
\end{itemize}

We show the zero-shot performance of \ours{} and baselines pretrained with the near-expert data (both in single-task and multi-task settings). We report L2 distance averaged over the states and goals sampled based on the above rules. Tables of run numbers and standard derivations can be found in~\Secref{sec:full_exp_result}.

\begin{figure}[t!]
    \centering
    \includegraphics[width=.98\textwidth]{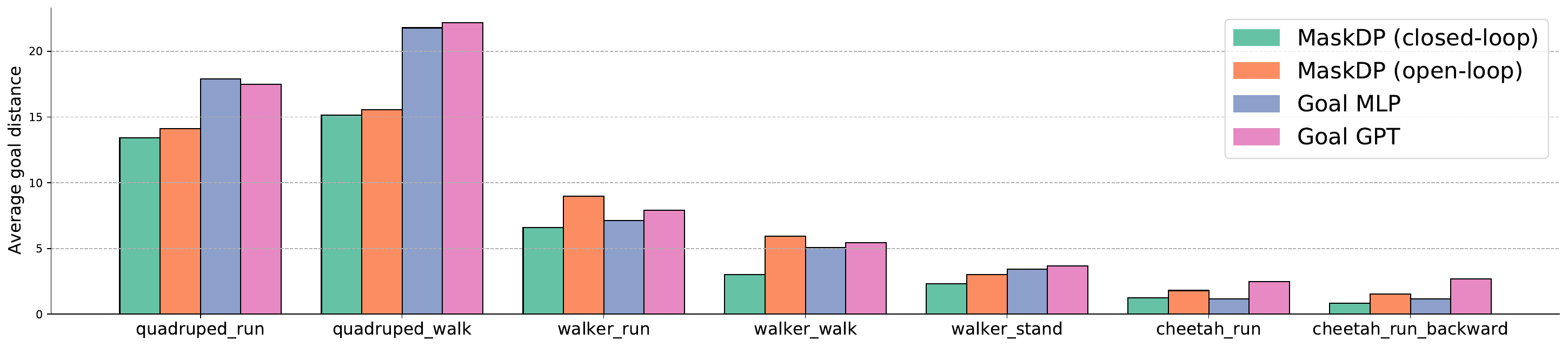}
    \caption{Single task pretraining followed by single goal reaching downstream task. \ours{} with closed-loop execution achieves the best performance on all the tasks, and get the most significant improvements in the Quadruped domain, which is higher dimensional.}
    \label{fig:single_task_single_goal}
\end{figure}

\Figref{fig:single_task_single_goal} show the results of reaching single goal. The y-axis is the L2 distance (the lower the better). We observe that both \ours{} (open-loop) and \ours{} (closed-loop) outperform Goal-GPT and Goal-MLP. Despite Goal-GPT being a natural formulation for goal reaching, \ours{} reaches a lower distance to the goal. We attribute the effectiveness to learning a better understanding of the forward and inverse dynamics implicitly.
We also observe that the advantages of \ours{} are even more significant in higher dimensional environments, such as Quadruped.

\begin{figure}[t!]
    \centering
    \includegraphics[width=\textwidth]{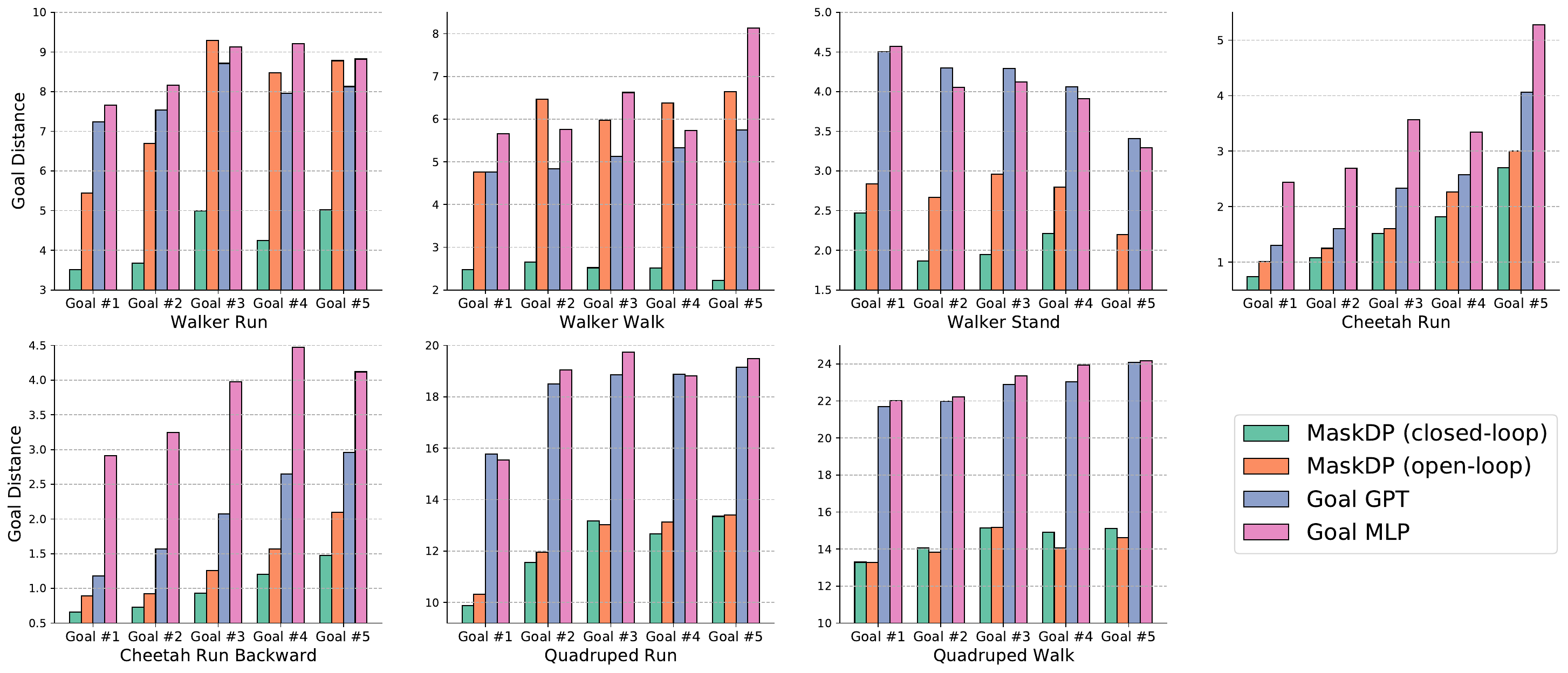}
    \caption{Single task pretraining followed by multiple goals reaching downstream task. \ours{} achieves significant improvement on all the tasks with better flexibility in sequential goal reaching.}
    \label{fig:single_task_multiple_goal}
\end{figure}

For the more challenging multi-goal reaching task, \ours{} has a significant advantage in flexibility: we can just provide the goals at specific time budgets with interleaved masks and get an executable plan; however, for Goal-MLP and Goal-GPT, we have to change goals at certain timesteps to fulfill future multiple goals. 
As shown in~\Figref{fig:single_task_multiple_goal}, \ours{} outperforms both goal-GPT and BC by a large margin. In~\Figref{fig:foresight}, we showed that having "foresight" about future goals can help the agent to generate a better plan.

\begin{figure}[!htbp]
    \centering
    \includegraphics[width=.98\textwidth]{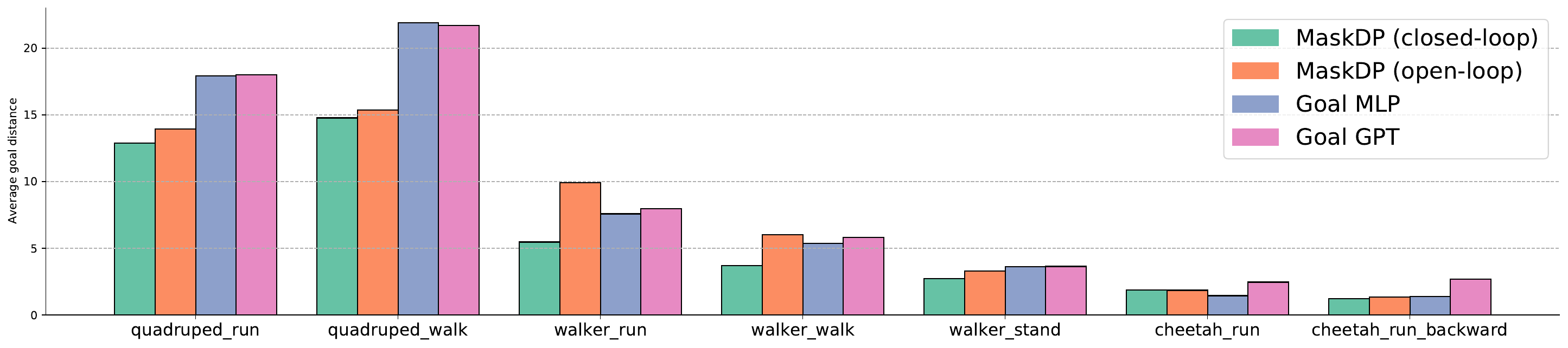}
    \caption{Multiple tasks pretraining followed by single goal reaching downstream task, where \ours{} with closed-loop execution works the best, especially in the Quadruped domain.}
    \label{fig:multiple_task_single_goal}
\end{figure}

\begin{figure}[!htbp]
    \centering
    \includegraphics[width=\textwidth]{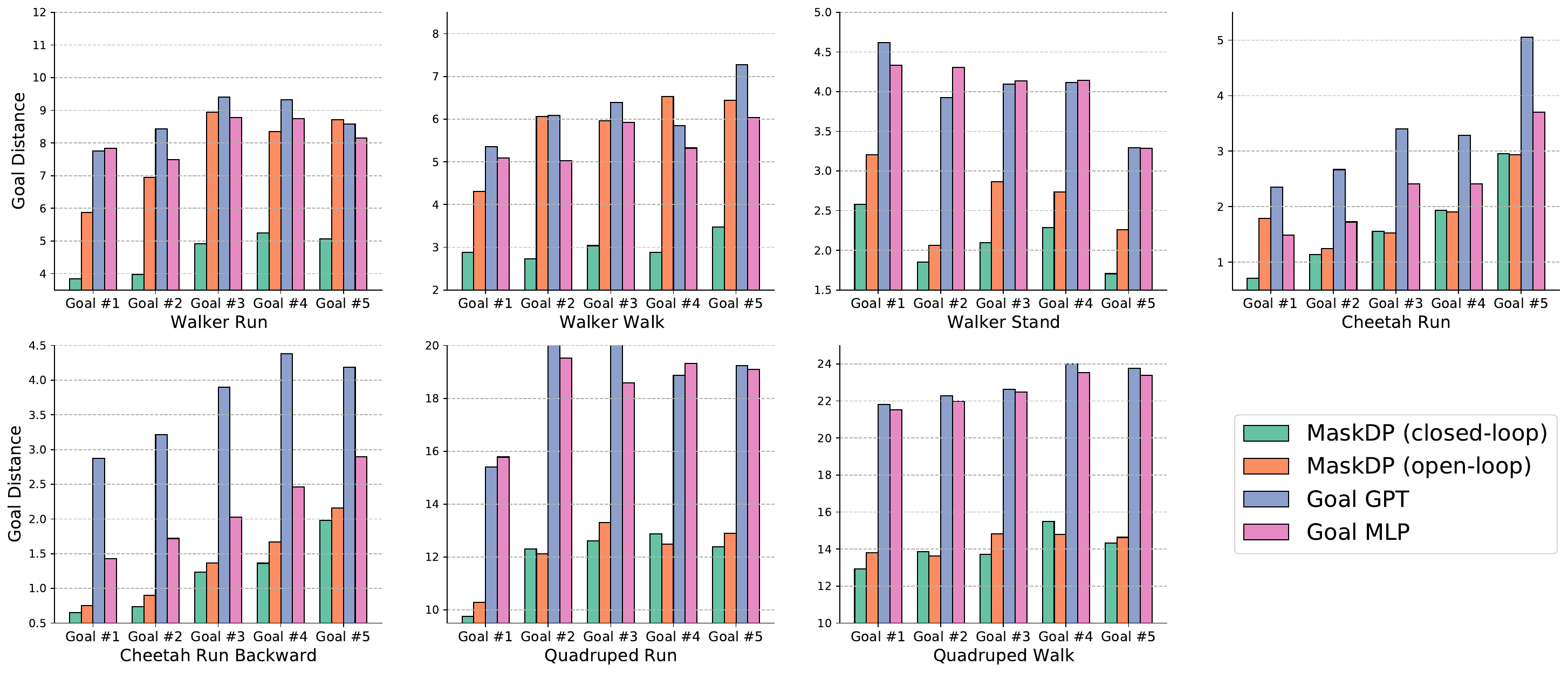}
    \caption{Multiple task pretraining followed by multiple goals reaching downstream task. 
    }
    \label{fig:multiple_task_multiple_goal}
\end{figure}

We can get similar conclusions from the multi-task pretrained models in~\Figref{fig:multiple_task_single_goal} and~\Figref{fig:multiple_task_multiple_goal}, where our method consistently works well on all domains, with the most visible advantage in multi-goal reaching setup.

\begin{figure}[!thbp]
    \centering
    \begin{tabular}{cc}
        \includegraphics[width=.98\textwidth]{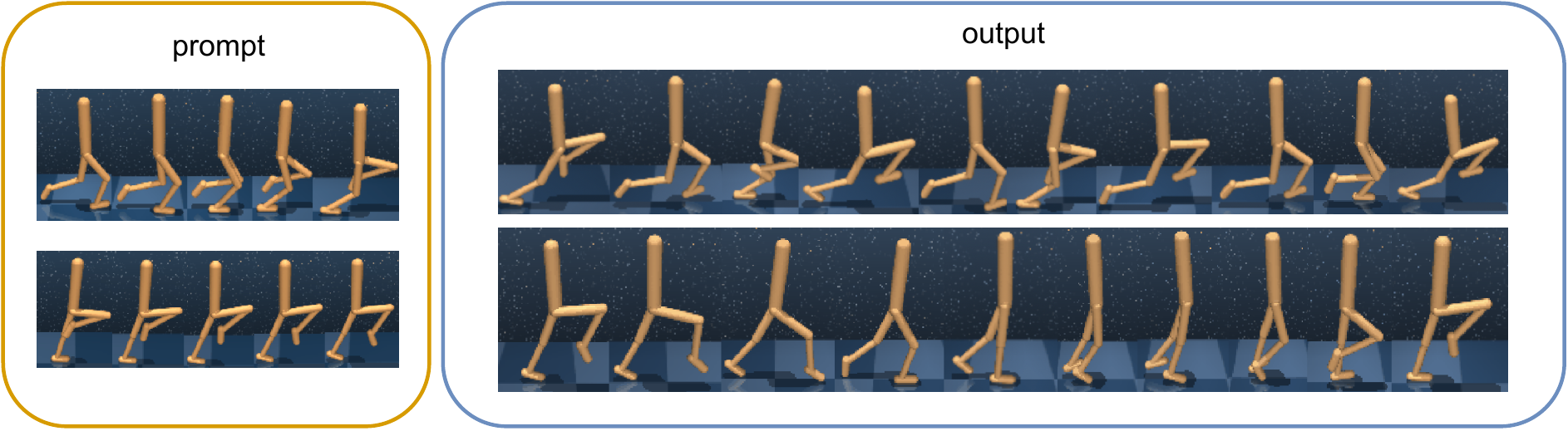} \\
    \end{tabular}
    \caption{Qualitative results on for skill prompting in the Walker domain. Given 5 initial states, the model learns to forecast future trajectories as in the expert-level behaviour.
    }
    \label{fig:vis-prompt}
\end{figure}

\subsubsection{Skill Prompting}
We are interested in the learned behavior of pretrained models. We use prompting, which has become popular in analyzing models ever since GPT~\citep{brown2020language}. To do so, we give the agent a short state-action segment randomly cropped from an expert trajectory, set the agent to the last state of the segment, and let the model continue to generate consecutive behaviors. We evaluate the quality of the generated sequence by comparing its obtained rewards with the rollout of a skilled expert. 

\begin{figure}[t!]
    \centering
    \begin{tabular}{cc}
        \includegraphics[width=.98\textwidth]{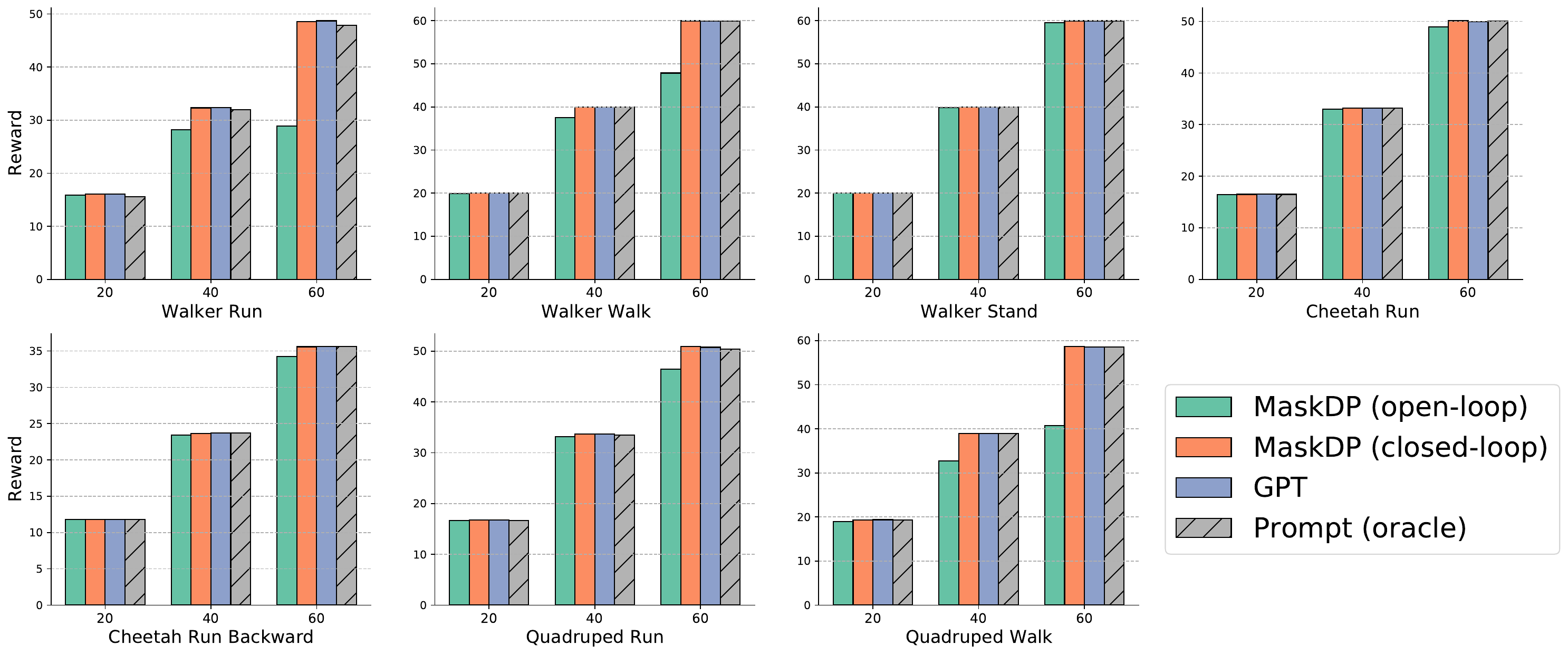} \\
    \end{tabular}
    \caption{Quantitative results on learned behaviors using prompt. Both MaskDP and GPT can match or even slightly surpass the expert-level performance (right grey bar) in trajectory forecasting. 
    }
    \label{fig:prompt}
\end{figure}

To be specific, we prompt the multi-task model trained with expert data and sample a 5-timestep state-action segment from $T \in [100, 900]$, where the agent can be walking/running at low or high speeds. We prompt the model with this short segment and let the model generate rollouts for 20, 40, 60 timesteps. We provide both qualitative results and quantitative results in~\Figref{fig:prompt} and~\Figref{fig:vis-prompt} respectively. We can see in~\Figref{fig:prompt}, both our method and GPT can match the expert return. It shows that our method can perform as well as autoregressive model in generation task. 

\subsubsection{Offline Reinforcement Learning}
\label{sec: rl}
\paragraph{Evaluation}  We provide a 2M buffer of the data collected by Proto-RL~\citep{yarats2021reinforcement} as in ExoRL~\citep{yarats2022dont} does, where the overall return of the data is quite low and thus the BC-based method cannot work well. ExoRL~\citep{yarats2022dont} simply shows that an offline TD3 agent works the best on diverse low-return offline data.

\begin{figure}[t!]
    \centering
    \includegraphics[width=\textwidth]{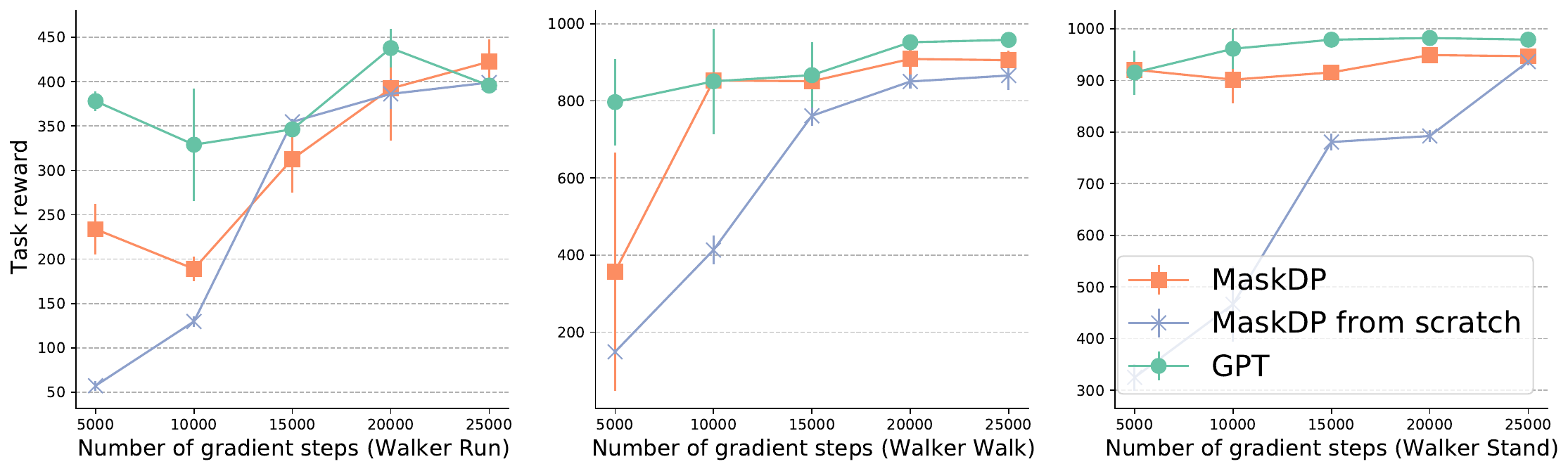}
    \caption{Offline RL results on Walker domain. The result of \ours{} matches the GPT-style pretraining performance, and are all comparable to the SoTA in ExoRL~\citep{yarats2022dont}.}
    \label{fig:rl_finetune}
\end{figure}

We can modify \ours{} to this setting by adding additional actor and critic heads on top of the pretrained encoder, and performing RL training. We evaluate the efficiency of the pretrained model by its return after certain TD gradient steps. The results are shown in~\Figref{fig:rl_finetune} averaged over 3 seeds. We observe \ours{} is capable of adapting to downstream tasks quickly, outperforms training from scratch, and achieves similar results as the GPT baseline. Note that in this setting, we need to replace the bidirectional attention mask with the causal attention mask, so there is a larger gap between pretraining and downstream tasks finetuning compared with GPT, which is trained with causal masking. Note that MaskDP from scratch is almost the same as GPT from scratch (both with causal masking). From~\Figref{fig:rl_finetune}, both \ours{} and GPT can match the best result in ExoRL~\citep{yarats2022dont} from their offline TD3 agent, where BC-based method cannot successfully solve this task.

\subsection{Analysis}
\label{sec:exp_analysis}

\begin{wrapfigure}[12]{r}{.675\textwidth}
    \vspace{-4mm}
    \centering
    \includegraphics[width=.675\textwidth]{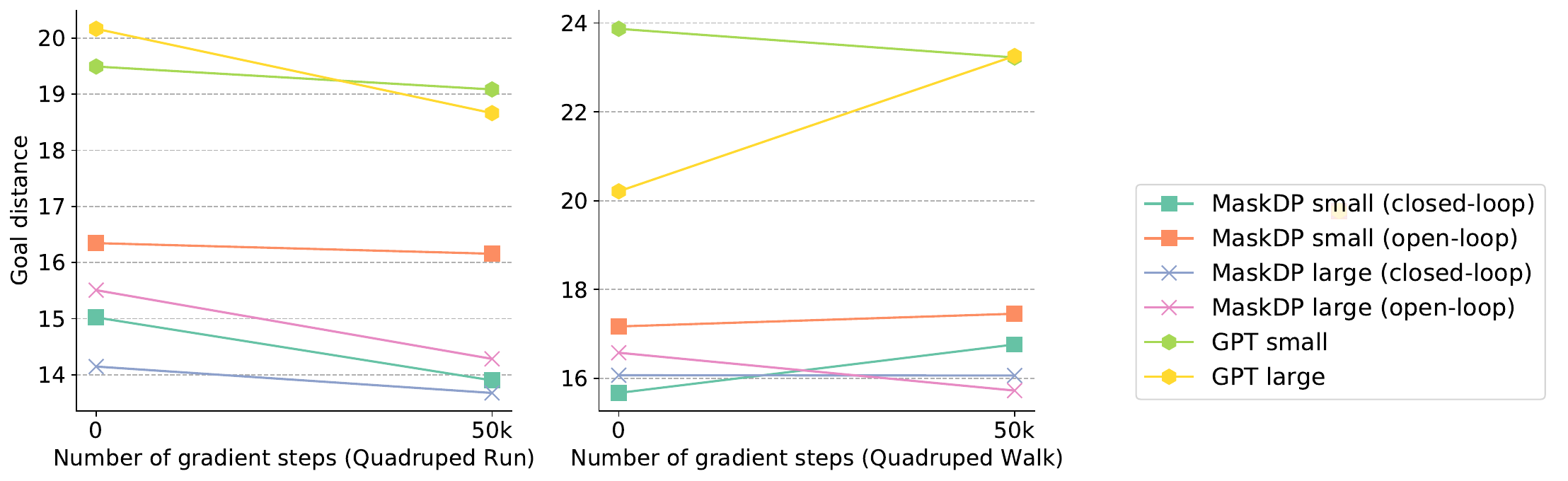}
    \caption{Model scalability on two tasks. X-axis represents number of gradient steps. With \ours{} pretraining, larger models outperform smaller models. 
    }
    \label{fig:scaling}
\end{wrapfigure}
\textbf{Model scalability.}
\label{sec:scalability}
We also pretrain our agent and baselines on the diverse "mixed" dataset. We compare the smaller version of \ours{} and Goal-GPT\footnote{Both use 3 attention layers.} on the Quadruped goal reaching problem as shown in~\Figref{fig:scaling}. The x-axis is the finetuning gradient steps on the expert dataset, and y-axis is the L2 distance to the goal (the lower the better). We found for both zero-shot evaluation and finetuning, our model's performance improves when model size is enlarged, whereas for Goal-GPT the performance gain is not obvious.

\begin{figure}[t!]
    \centering
    \includegraphics[width=.95\textwidth]{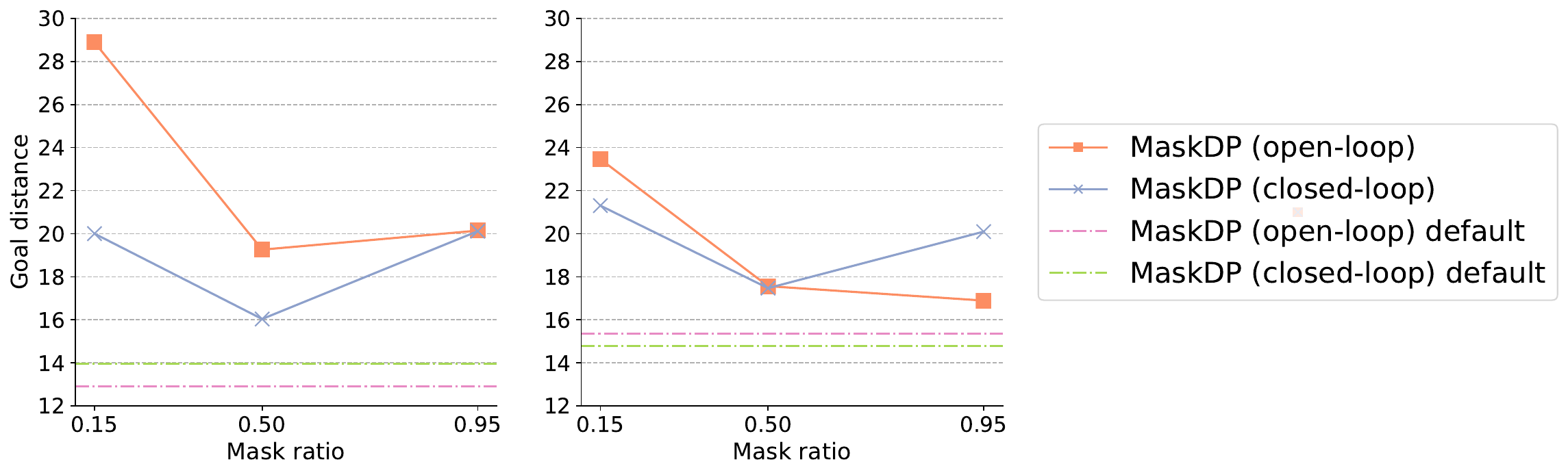}
    \caption{Mask ratio ablation. We compare our multiple ratio pretrained model with models trained with fixed ratios, where our masking strategy can achieve much better performance.}
    \vspace{-1.0\intextsep}
    \label{fig:mask_ratio_ablation}
\end{figure}

\textbf{Mask ratio ablation.}
\Figref{fig:mask_ratio_ablation} shows the inﬂuence of the masking ratio. 
With a fixed mask ratio, we observe that an extremely high mask ratio (95\%) generally does not work well and the typical mask ratio (15\%) used in BERT seems to perform much worse than others. A middle mask ratio 50\% performs reasonably well, despite still being surprisingly high, similiar to the observations in MAE. However, our mixed mask ratio strategy strictly outperforms all the above options.

\begin{wrapfigure}[10]{r}{.55\textwidth}
    \vspace{-4mm}
    \centering
    \includegraphics[width=.55\textwidth]{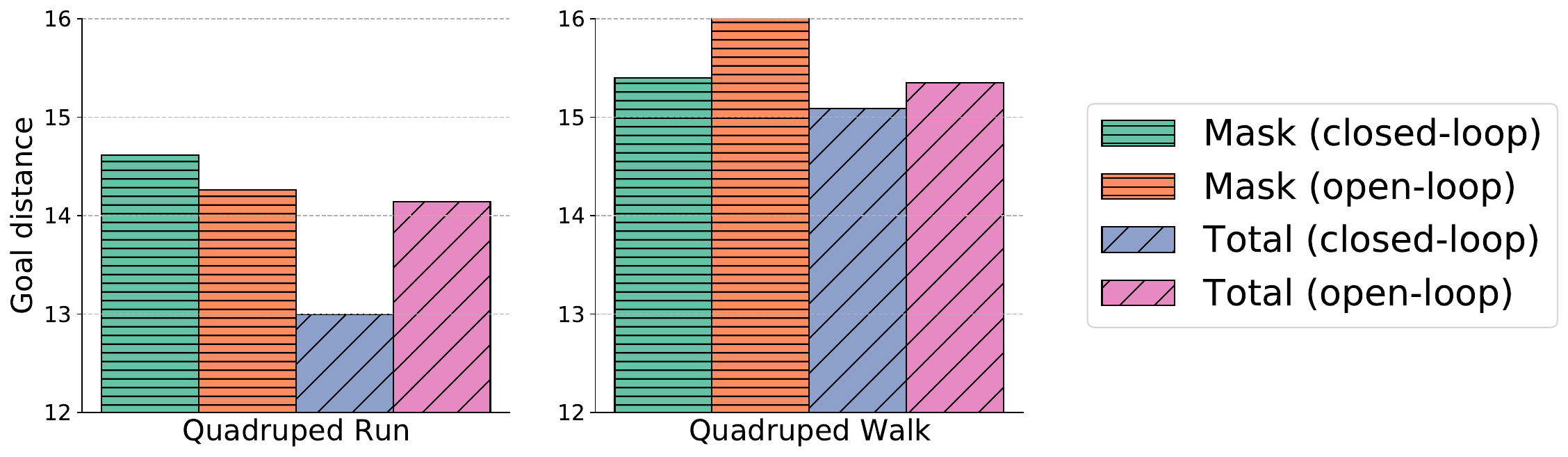}
    \caption{Masked loss and total loss ablation on 20k pretraining gradient steps. The model trained with total loss converges faster than the one trained with masked loss.}
    \label{fig:loss_ablation}
\end{wrapfigure}
\textbf{Predicting unmasked tokens ablation.}
We also compare the model trained with mask loss vs. total loss. As shown in~\Figref{fig:loss_ablation}, empirically we do not find mask loss has more advantages than total loss, even on the relatively clean expert dataset, it converges slower than using total loss. For the results on diverse data, please refer to~\Secref{sec:full_exp_result}.

\section{Conclusion}
This paper presents masked decision prediction (\ours{}), a simple and scalable self-supervised method for reinforcement learning (RL) inspired by current large language and vision models.
\ours{} is capable of learning scalable and generalizable agents for reinforcement learning that can learn from diverse-quality data sources and infer tasks in goal-reaching and skill-execution settings. 
Through our empirical study we ﬁnd MaskDP models outperform past work in zero-shot goal reaching and transfer well to downstream RL tasks, performing competitively with prior pretraining and training from scratch methods. 

\paragraph{Limitations and Future Work}
Computer vision and NLP domains have shown that the true promise of masking architectures lies with their ability to ingest diverse, fully unsupervised data. We study how \ours{} performs when trained without access to expert-level data, and evaluated on unseen proprioceptive states. In the future, we can extend our method to pixel inputs, and pretrain the model to adapt to far different downstream tasks.

The architecture used in \ours{} closely resembles a model-based method, as states are predicted sequentially from actions. In this paper, we use use the predicted next actions directly as this is the simplest and fastest approach. However, it is straightforward to extend \ours{} to plan through our learned model and compare against related baselines.

\paragraph{Societal Impact}

This is an algorithm for training agents in the style of recent large-scale CV and NLP models. While we do not anticipate particular social risks from our method, as algorithms become capable of ingesting large-scale, in-the-wild data it is important to ensure the dataset does not reinforce undesirable biases or promote harmful behaviors.

\section*{Acknowledgment}
This work was partially supported by Office of Naval Research (under grant N00014-21-1-2769 and grant N00014-22-1-2121) and National Science Foundation (under grant NSF NRI \#2024675). 
We would like to thank Olivia Watkins, Yuqing Du, Younggyo Seo, Xingyu Lin and Danijar Hafner for giving constructive comments. We would also like to thank anonymous reviewers for their helpful feedback.


\bibliography{ms}
\bibliographystyle{abbrvnat}

\newpage
\input{appendix}

\end{document}

%% file: math_commands.tex
\usepackage{amsmath,amsfonts,bm}

\def\Figref#1{Figure~\ref{#1}}

\def\Secref#1{Section~\ref{#1}}

\def\eqref#1{equation~(\ref{#1})}

\def\1{\bm{1}}

\DeclareMathAlphabet{\mathsfit}{\encodingdefault}{\sfdefault}{m}{sl}
\SetMathAlphabet{\mathsfit}{bold}{\encodingdefault}{\sfdefault}{bx}{n}



%% file: appendix.tex
\appendix
\section{Experimental Details}
\label{sec:exp_detail}

\subsection{Environments and Tasks}
We provide the details about the environments and tasks used in our experiments in Table~\ref{tab:env}.

\begin{table}[htbp!]
\begin{center}
\begin{tabular}{|c|cc|ccc|cc|}
\hline
Domain     & \multicolumn{2}{c|}{Quadruped} & \multicolumn{3}{c|}{Walker} & \multicolumn{2}{c|}{Cheetah} \\ \hline
Task       & run           & walk           & run    & walk    & stand    & run      & run\_backward      \\ \hline
State dim  & \multicolumn{2}{c|}{79}        & \multicolumn{3}{c|}{18}     & \multicolumn{2}{c|}{24}      \\ \hline
Action dim & \multicolumn{2}{c|}{12}        & \multicolumn{3}{c|}{6}      & \multicolumn{2}{c|}{6}       \\ \hline
\end{tabular}
\end{center}
\caption{Environments and tasks from the DeepMind control suite~\cite{tassa2018deepmind} used in our experiments.}
\label{tab:env}
\vskip 0.1in
\end{table}

\subsection{Datasets}
\paragraph{Near-Expert Dataset} We use TD3 agent~\cite{fujimoto2018addressing} trained with 1M steps in the above tasks. We then freeze its parameters and rollout its policy with Gaussian noise $N(0, 0.2)$ in every dimension of the action space. For each task, we collect 4M steps of experience (4K episodes in total) using this pretrained TD3 agent.  
\paragraph{Mixed Dataset} Mixed dataset consists of the following data from various RL agents:
\begin{itemize}
    \item Near-expert data: Same as the above near-expert dataset, but we only include 2M steps experience (2K episodes in total) for each task.
    \item Unsupervised data: We use an unsupervised RL algorithm, Proto-RL~\cite{yarats2021reinforcement}, to collect diverse unsupervised data. We train the agent for 10M steps in each domain and record all the 10M steps (10K episodes).
    \item Semi-supervised data: We train a TD3 agent to optimize the sum of Proto-RL~\cite{yarats2021reinforcement} intrinsic reward and extrinsic reward. The agent is trained for 2M steps in each task and we record all the 2M steps (2K episodes) for each task.
\end{itemize}

\subsection{Hyperparameters}
We provide more details about the hyperparameters and other settings of model training and evaluation in Table~\ref{tab:hyperparam}.

\begin{table}[htbp!]
\vskip 0.15in
\begin{center}
\begin{small}
\begin{tabular}{ll}
\toprule
\textbf{\ours{}} & \textbf{Value}  \\
\midrule
\# Context length & 64 \\
\# Encoder layer & $3$ \\ 
\# Decoder layer & $2$ \\
\# Attention head & $4$ \\ 
\# Hidden dimension & $256$ \\
Mask ratio & [0.15, 0.35, 0.55, 0.75, 0.95] \\
\midrule
\textbf{GPT/Goal-GPT} & \textbf{Value}  \\
\midrule
\# Context length & 64 \\
\# Attention layer & $5$ \\ 
\# Attention head & $4$ \\ 
\# Hidden dimension & $256$ \\
\midrule
\textbf{Goal-MLP} & \textbf{Value}  \\
\midrule
\# Context length & 64 \\
\# Linear layer & $5$ \\ 
\# Hidden dimension & $1024$ \\
\midrule
\textbf{Training} & \textbf{Value}  \\
\midrule
Optimizer    & Adam  \\ 
$(\beta_1,\beta_2)$  & $(.9,.999)$  \\
Learning rate  & $1e^{-4}$ \\
Batch size    & $384$   \\ 
\# Gradient step & $400000$ \\
\midrule
\textbf{Evaluation} & \textbf{Value}  \\
\midrule
\# seed & $3$ \\
\# Goals (single-goal reaching) per seed  & $300$  \\ 
\# Goals (multi-goal reaching) per seed  & $100 \times 5$  \\
Prompt context length & $5$ \\
Discount (for RL) & $0.99$ \\ 
Replay buffer size (for RL) & $2M$ \\ 
\bottomrule
\end{tabular}
\end{small}
\caption{Hyperparameters used for model training and evaluation.}
\label{tab:hyperparam}
\end{center}
\vskip 0.1in
\end{table}

\subsection{Training Details}
\paragraph{Goal-MLP Training} We adapt the training of Goal-MLP to make it learn to reach goals with varying time budgets. Given a state-action sequence ($s_t$, $a_t$, $s_{t+1}$, ..., $s_{t+m}$), Goal-MLP randomly sample two states $s_{i}$ and $s_{j}$ as starting state and goal, (where $t \leq i < j \leq t+m$), and predicts the action $a_i$. 
\paragraph{Goal-GPT Training} Given a state-action sequence ($s_t$, $a_t$, $s_{t+1}$, ..., $s_{t+m}$), Goal-GPT treats $g = s_{t+m}$ as goal. Every state $s_{i}$ (where $t \leq i < t+m$) is concatenated with $g$. Then Goal-GPT predicts the action sequence $a_t$, ...  $a_{t+m-1}$ from the state-goal sequence $(s_t, g)$, ...  $(s_{t+m-1}, g)$ by passing through causal self-attention layers. 
In this way, all the goal-reaching baselines are pre-trained to reach goals in various timesteps.
\paragraph{GPT Training} Given a state-action sequence ($s_t$, $a_t$, $s_{t+1}$, ..., $s_{t+m}$), GPT predicts the next token (state or action) conditions on previous token sequence, $i.e.$, predicting $s_j$ ($j > t$) from $s_t$, $a_t$, ... $s_{j-1}$, $a_{j-1}$.

\subsection{Compute Resources}
\ours{} is designed to be accessible to the RL research community. The whole pipeline, including data collection, pretraining, and finetuning, only requires a single GPU. 
All experiments were run on GPU clusters with 8 NVIDIA TITAN Xp. 
The pretraining takes 6-8 hours for 400k gradient steps on the collected datasets using a single GPU.

\section{{Additional Experimental Study}}
\subsection{Dataset Quality}
\label{sec:mixv2}
MaskDP has no assumption about the pretraining dataset. To show it doesn't rely on the expert data, we reconstruct another highly diverse dataset called \textbf{mixed-v2}, which contains:
\begin{itemize}
    \item unsupervised data: we train a TD3~\citep{fujimoto2018addressing} agent to maximize Proto-RL~\citep{yarats2021reinforcement} intrinsic reward, and store its 10M replay buffer on each domain.
    \item semi-supervised data: we train a TD3 agent to maximize the sum of extrinsic reward and the Proto-RL intrinsic reward, and store its 2M replay buffer on each task.
    \item supervised data: we train a TD3 agent to maximize extrinsic reward and store its 2M replay buffer on each task.
\end{itemize}
The \textbf{mixed-v2} dataset is more diverse, as it replaces the near-expert data with TD3 training samples, which are more suboptimal and noisy. After pretraining using \textbf{mixed-v2}, we evaluate its performance on unseen state-goal pairs from \textbf{near-expert} dataset (dataset in the main paper). So the pretraining and evaluation datasets are in different distributions.

\begin{figure}[!htbp]
    \centering
    \includegraphics[width=\textwidth]{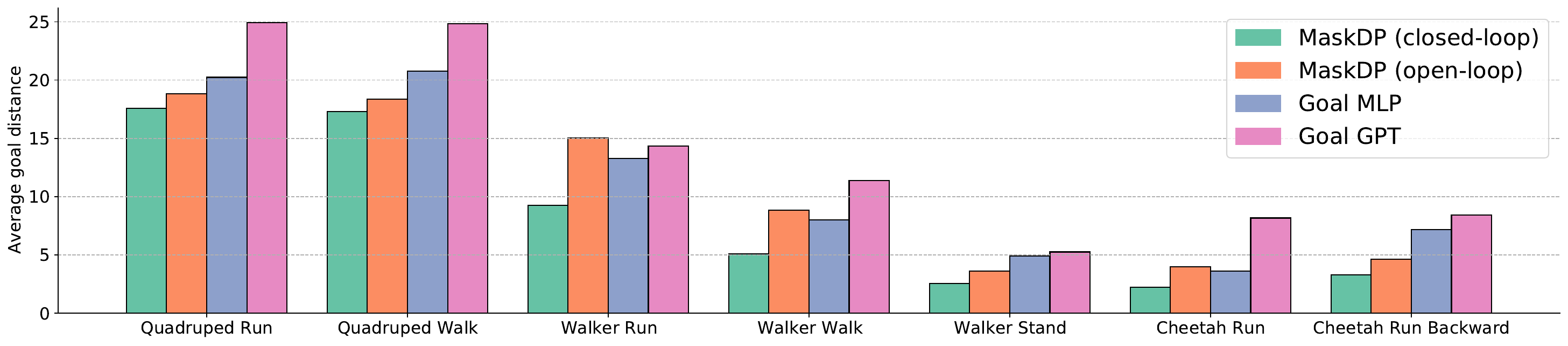}
    \caption{The single goal reaching results on \textbf{near-expert} goal reaching, after pretraining MaskDP on \textbf{mixed-v2} dataset.
    }
    \label{fig:single-new}
\end{figure}

\begin{figure}[!htbp]
    \centering
    \includegraphics[width=\textwidth]{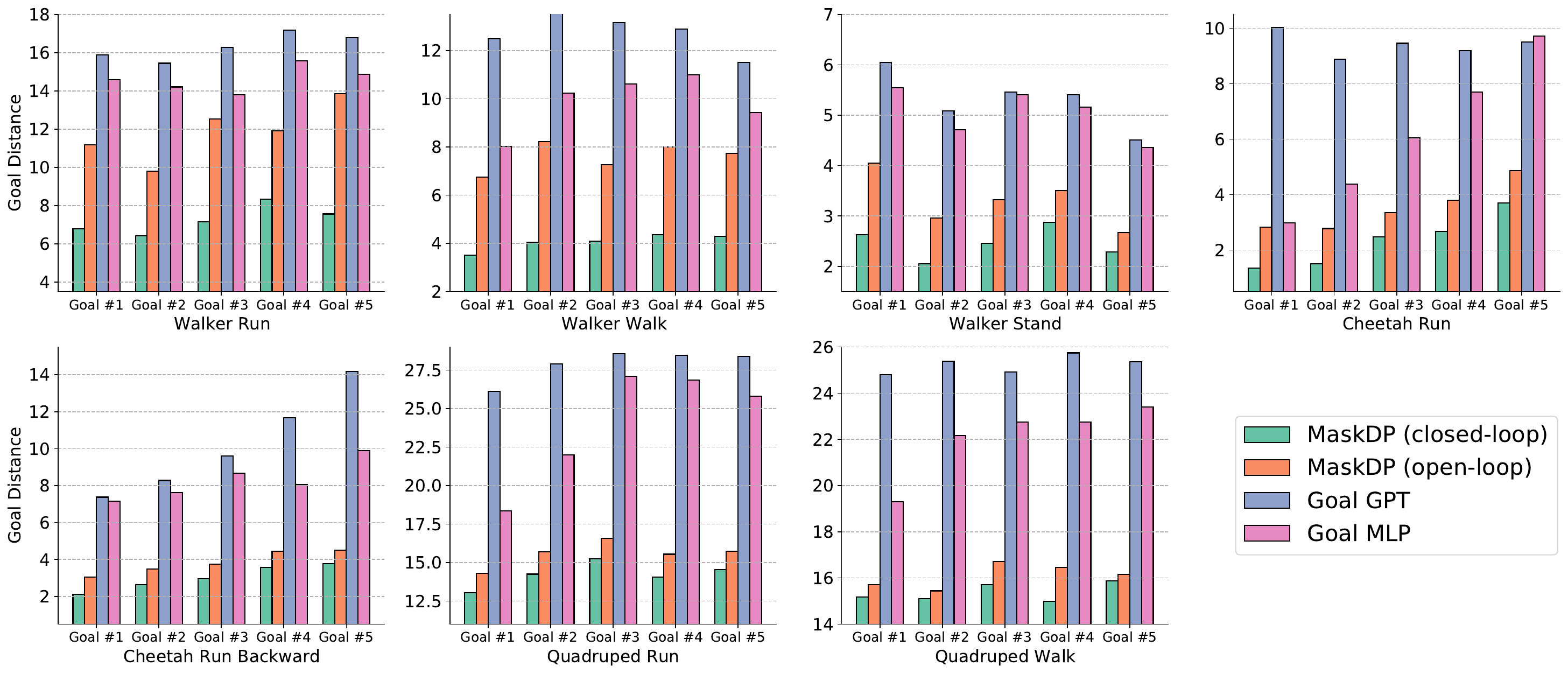}
    \caption{The multiple goal reaching results on \textbf{near-expert} goal reaching, after pretraining MaskDP on \textbf{mixed-v2} dataset.
    }
    \label{fig:multi-new}
\end{figure}

In Figure~\ref{fig:single-new} and Figure~\ref{fig:multi-new}, we find our model consistently outperforms baselines on all the domains. Compared with Figure~\ref{fig:multiple_task_single_goal} and Figure~\ref{fig:multiple_task_multiple_goal}, it has more advantages when dataset is noisy, as BC-based methods highly rely on the dataset quality. 

\subsection{Foresight Helps Multi-goal Reaching}
We add ablation about whether to provide multiple future goals to the agent for multi-goal reaching. In contrast, We can also give the agent an individual goal at one time, and switch to a different goal when the budget is exhausted. 

\begin{figure}[!htbp]
    \centering
    \includegraphics[width=\textwidth]{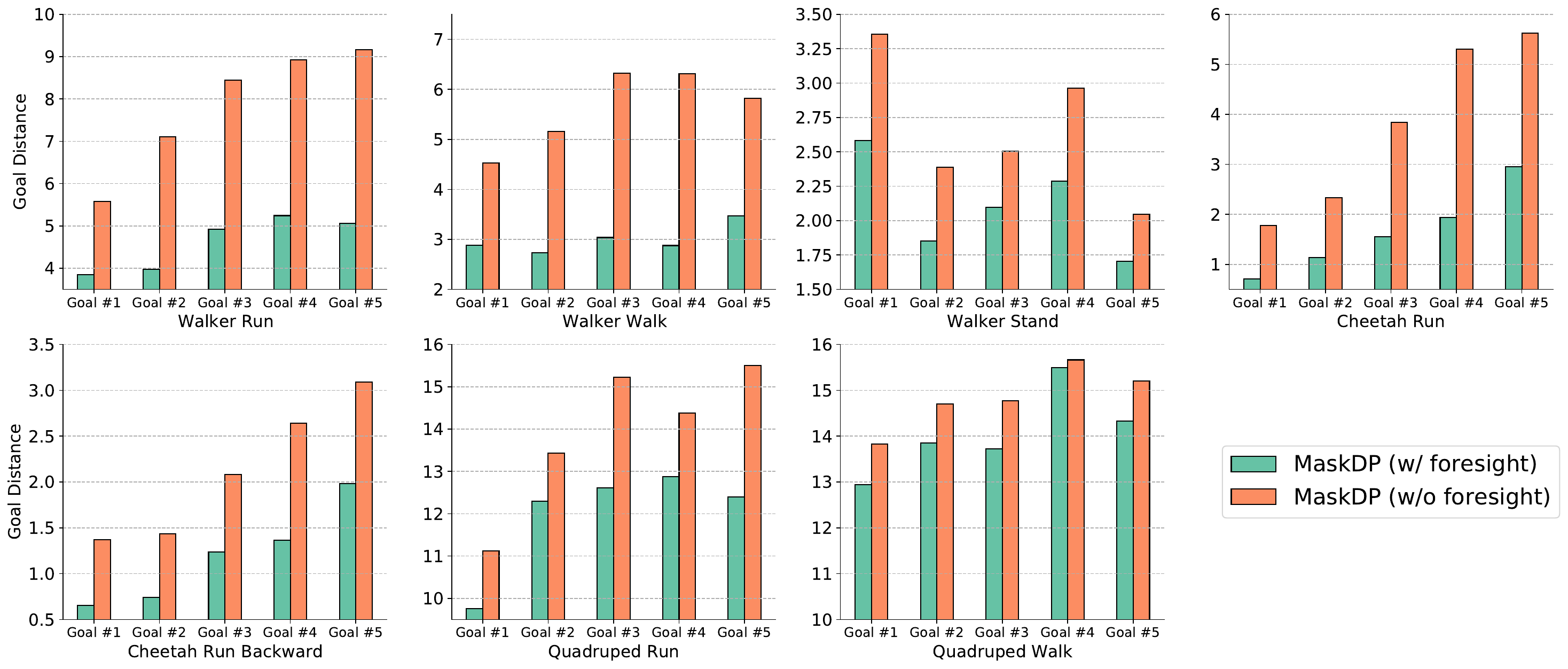}
    \caption{We test the closed-loop performance of MaskDP to understand whether the visibility of future goals can improve the performance. We found that on all the domains, MaskDP with foresight performs better. 
    }
    \label{fig:foresight}
\end{figure}
\subsection{Trajectory Length Affects Generation Quality}
\begin{figure}[!htbp]
    \centering
    \includegraphics[width=\textwidth]{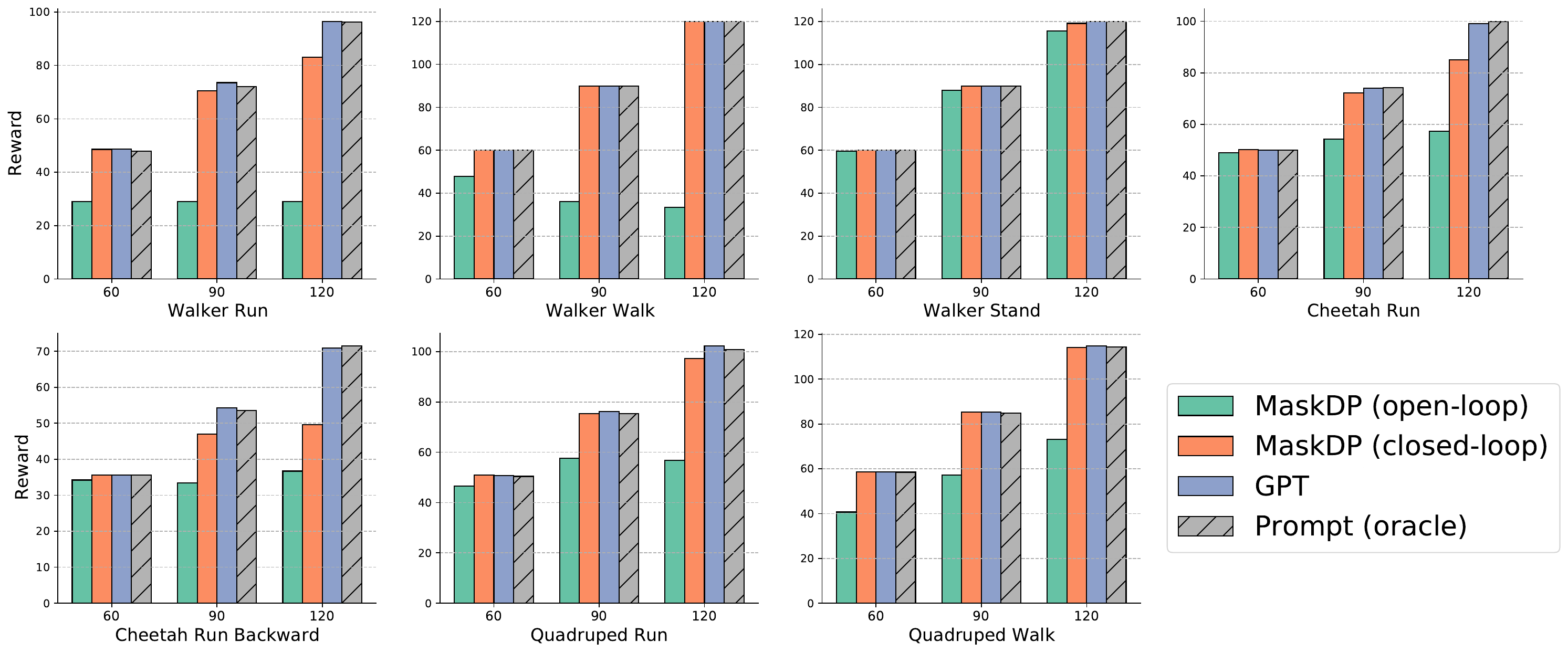}
    \caption{Skill prompting performance for longer rollouts. 
    }
    \label{fig:prompt-len}
\end{figure}

As in \cite{he2021masked,DEIT}, we also use sinusoidal positional embedding and perform linear interpolation when the trajectory is longer than the training time. Figure~\ref{fig:prompt-len} shows the results when we execute the agent for 60, 90, and 120 steps with 5 context tokens, where the training trajectory length is 64.
We found on most environments, closed-loop MaskDP can achieve similar performance with GPT and the expert return (the gray bar), except for Cheetah tasks. For longer trajectories, the mask ratio can be extremely low at the beginning, which can cause some bad initial behavior. Meanwhile, GPT can perform stably well as it's not conditioned on masked inputs. 
%
\subsection{Additional Domain: Jaco}
\begin{figure}[!htbp]
    \centering
    \includegraphics[width=\textwidth]{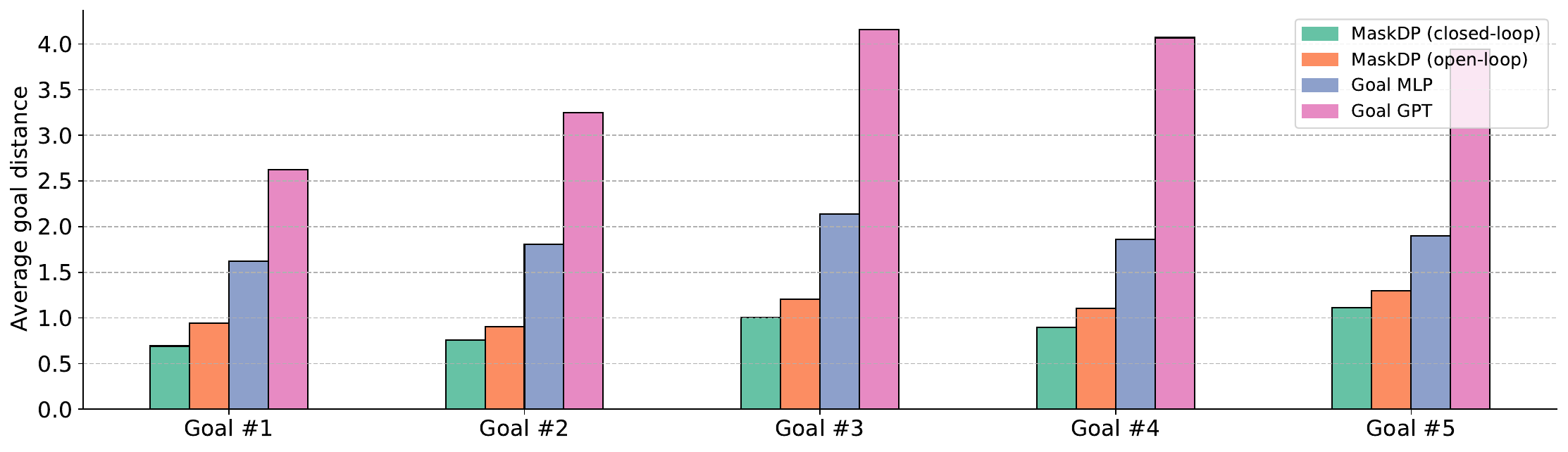}
    \caption{Multi-goal reaching results on Jaco reaching task. MaskDP outperforms baselines on this domain.
    }
    \label{fig: Jaco}
\end{figure}

We also add a Jaco arm reaching task in the robotics domain. The training and evaluation both follow \Secref{sec:mixv2} As shown in Figure~\ref{fig: Jaco}, MaskDP still outperforms baselines on this task.

\section{Full Experimental Results}
\label{sec:full_exp_result}
\subsection{Single-goal Reaching Results}
We provide the single-goal reaching results in Table~\ref{tab:st-single-goal} and Table~\ref{tab:mt-single-goal} for single-task and multi-task respectively.
\begin{table}[htbp!]
\begin{tabular}{|c|c|c|c|c|c|}
\hline
Domain                     & Task         & Goal-MLP         & Goal-GPT         & Ours (open-loop) & Ours (closed-loop) \\ \hline
\multirow{2}{*}{Quadruped} & run          & 17.832$\pm$0.321 & 18.313$\pm$0.171 & 13.753$\pm$0.255 & 12.912$\pm$0.018   \\
                           & walk         & 22.965$\pm$0.077 & 23.051$\pm$0.055 & 15.456$\pm$0.176 & 15.116$\pm$0.396   \\ \hline
\multirow{3}{*}{Walker}    & run          & 8.15$\pm$0.080   & 9.197$\pm$0.012  & 9.565$\pm$0.157  & 7.789$\pm$0.220    \\
                           & walk         & 5.111$\pm$0.118  & 6.029$\pm$0.415  & 5.822$\pm$0.281  & 2.63$\pm$0.158     \\
                           & stand        & 3.979$\pm$0.293  & 4.108$\pm$0.317  & 3.242$\pm$0.1527 & 2.393$\pm$0.076    \\ \hline
\multirow{2}{*}{Cheetah}   & run          & 1.377$\pm$0.037  & 2.878$\pm$0.057  & 2.234$\pm$0.145  & 1.385$\pm$0.122    \\
                           & run backward & 1.447$\pm$0.136  & 3.011$\pm$0.095  & 1.752$\pm$0.052  & 0.866$\pm$0.065    \\ \hline
\end{tabular}
\caption{single-task single-goal reaching results.}
\label{tab:st-single-goal}
\end{table}

\begin{table}[htbp!]
\begin{tabular}{|c|c|c|c|c|c|}
\hline
Domain                     & Task         & Goal-MLP          & Goal-GPT          & Ours (open-loop)  & Ours (closed-loop) \\ \hline
\multirow{2}{*}{Quadruped} & run          & 17.825$\pm$0.218 & 18.224$\pm$0.551 & 13.214$\pm$0.230 & 12.926$\pm$0.382  \\
                           & walk         & 22.977$\pm$0.484 & 23.361$\pm$0.201 & 14.892$\pm$0.140 & 15.428$\pm$0.203  \\ \hline
\multirow{3}{*}{Walker}    & run          & 8.600$\pm$0.095    & 8.977$\pm$0.181  & 10.242$\pm$0.149 & 6.233$\pm$0.149   \\
                           & walk         & 5.550$\pm$0.170   & 6.083$\pm$0.392  & 6.584$\pm$0.487  & 4.042$\pm$0.148   \\
                           & stand        & 4.096$\pm$0.015  & 4.137$\pm$0.408  & 3.422$\pm$0.161  & 2.248$\pm$0.026   \\ \hline
\multirow{2}{*}{Cheetah}   & run          & 1.634$\pm$0.042   & 2.953$\pm$0.085   & 1.924$\pm$0.099  & 1.939$\pm$0.125   \\
                           & run backward & 1.694$\pm$0.061  & 2.980$\pm$0.076   & 1.378$\pm$0.055   & 1.395$\pm$0.011   \\ \hline
\end{tabular}
\caption{Multi-task single-goal reaching results.}
\label{tab:mt-single-goal}
\end{table}

\subsection{Multi-goal Reaching Results}
We provide multi-goal reaching results for multi-task pretrained models in Table~\ref{tab:first}, Table~\ref{tab:second}, Table~\ref{tab:third}, Table~\ref{tab:fourth} and Table~\ref{tab:fifth}. 
\begin{table}[htbp!]
\begin{tabular}{|c|c|c|c|c|c|}
\hline
Domain                     & Task         & Goal-MLP         & Goal-GPT         & Ours (open-loop) & Ours (closed-loop) \\ \hline
\multirow{2}{*}{Quadruped} & run          & 15.644$\pm$0.193 & 15.925$\pm$0.726 & 10.396$\pm$0.152 & 10.213$\pm$0.644   \\
                           & walk         & 21.963$\pm$0.241 & 22.114$\pm$0.445 & 13.413$\pm$0.545 & 12.99$\pm$0.073    \\ \hline
\multirow{3}{*}{Walker}    & run          & 7.562$\pm$0.385  & 7.588$\pm$0.243  & 5.981$\pm$0.006  & 4.032$\pm$0.265    \\
                           & walk         & 5.238$\pm$0.212  & 5.481$\pm$0.172  & 4.256$\pm$0.080  & 2.721$\pm$0.213    \\
                           & stand        & 4.295$\pm$0.050  & 4.483$\pm$0.189  & 2.998$\pm$0.271  & 2.293$\pm$0.426    \\ \hline
\multirow{2}{*}{Cheetah}   & run          & 1.381$\pm$0.151  & 2.342$\pm$0.009  & 0.995$\pm$0.132  & 0.738$\pm$0.041    \\
                           & run backward & 1.356$\pm$0.102  & 2.829$\pm$0.058  & 0.811$\pm$0.089  & 0.647$\pm$0.007    \\ \hline
\end{tabular}
\caption{Distance to the first goal in multi-task multi-goal reaching.}
\label{tab:first}
\end{table}

\begin{table}[htbp!]
\begin{tabular}{|c|c|c|c|c|c|}
\hline
Domain                     & Task         & Goal-MLP         & Goal-GPT         & Ours (open-loop) & Ours (closed-loop) \\ \hline
\multirow{2}{*}{Quadruped} & run          & 18.183$\pm$1.883 & 17.915$\pm$0.649 & 11.44$\pm$1.065  & 11.736$\pm$0.796   \\
                           & walk         & 22.628$\pm$1.006 & 23.225$\pm$1.325 & 13.986$\pm$0.515 & 14.487$\pm$0.937   \\ \hline
\multirow{3}{*}{Walker}    & run          & 8.161$\pm$0.939  & 8.98$\pm$0.775   & 7.165$\pm$0.305  & 4.398$\pm$0.599    \\
                           & walk         & 5.576$\pm$0.778  & 6.458$\pm$0.509  & 6.435$\pm$0.507  & 2.886$\pm$0.217    \\
                           & stand        & 4.389$\pm$0.119  & 4.013$\pm$0.131  & 2.566$\pm$0.508  & 2.045$\pm$0.282    \\ \hline
\multirow{2}{*}{Cheetah}   & run          & 1.814$\pm$0.126  & 2.75$\pm$0.115   & 1.207$\pm$0.056  & 1.163$\pm$0.032    \\
                           & run backward & 1.802$\pm$0.118  & 3.341$\pm$0.180  & 0.917$\pm$0.022  & 0.853$\pm$0.161    \\ \hline
\end{tabular}
\caption{Distance to the second goal in multi-task multi-goal reaching.}
\label{tab:second}
\end{table}

\begin{table}[htbp!]
\begin{tabular}{|c|c|c|c|c|c|}
\hline
Domain                     & Task         & Goal-MLP         & Goal-GPT         & Ours (open-loop) & Ours (closed-loop) \\ \hline
\multirow{2}{*}{Quadruped} & run          & 18.377$\pm$0.324 & 18.911$\pm$1.58  & 12.334$\pm$1.26  & 12.085$\pm$0.710   \\
                           & walk         & 22.787$\pm$0.269 & 23.907$\pm$0.946 & 14.7$\pm$0.166   & 14.069$\pm$0.498   \\ \hline
\multirow{3}{*}{Walker}    & run          & 9.05$\pm$0.3809  & 9.178$\pm$0.278  & 9.053$\pm$0.157  & 5.045$\pm$0.179    \\
                           & walk         & 6.127$\pm$0.268  & 6.603$\pm$0.489  & 5.895$\pm$0.097  & 3.113$\pm$0.106    \\
                           & stand        & 4.227$\pm$0.193  & 4.093$\pm$0.225  & 2.878$\pm$0.018  & 2.012$\pm$0.115    \\ \hline
\multirow{2}{*}{Cheetah}   & run          & 2.329$\pm$0.108  & 3.216$\pm$0.259  & 1.456$\pm$0.085  & 1.519$\pm$0.048    \\
                           & run backward & 2.11$\pm$0.114   & 3.807$\pm$0.131  & 1.291$\pm$0.102  & 1.216$\pm$0.026    \\ \hline
\end{tabular}
\caption{Distance to the third goal in multi-task multi-goal reaching.}
\label{tab:third}
\end{table}
\begin{table}[htbp!]
\begin{tabular}{|c|c|c|c|c|c|}
\hline
Domain                     & Task         & Goal-MLP         & Goal-GPT         & Ours (open-loop) & Ours (closed-loop) \\ \hline
\multirow{2}{*}{Quadruped} & run          & 19.564$\pm$0.142 & 19.227$\pm$0.156 & 12.92$\pm$0.52   & 12.89$\pm$0.030    \\
                           & walk         & 23.475$\pm$0.342 & 24.119$\pm$0.165 & 14.235$\pm$0.800 & 14.607$\pm$0.698   \\ \hline
\multirow{3}{*}{Walker}    & run          & 8.374$\pm$0.536  & 8.926$\pm$0.582  & 7.787$\pm$0.789  & 4.713$\pm$0.751    \\
                           & walk         & 5.548$\pm$0.282  & 5.988$\pm$0.193  & 6.248$\pm$0.402  & 2.884$\pm$0.007    \\
                           & stand        & 4.195$\pm$0.086  & 4.07$\pm$0.103   & 2.704$\pm$0.043  & 2.144$\pm$0.210    \\ \hline
\multirow{2}{*}{Cheetah}   & run          & 2.501$\pm$0.130  & 3.537$\pm$0.361  & 1.929$\pm$0.034  & 1.971$\pm$0.047    \\
                           & run backward & 2.491$\pm$0.047  & 4.145$\pm$0.330  & 1.825$\pm$0.217  & 1.48$\pm$0.166     \\ \hline
\end{tabular}
\caption{Distance to the fourth goal in multi-task multi-goal reaching.}
\label{tab:fourth}
\end{table}
\begin{table}[htbp!]
\begin{tabular}{|c|c|c|c|c|c|}
\hline
Domain                     & Task         & Goal-MLP         & Goal-GPT         & Ours (open-loop) & Ours (closed-loop) \\ \hline
\multirow{2}{*}{Quadruped} & run          & 18.749$\pm$0.796 & 19.083$\pm$1.376 & 13.057$\pm$0.202 & 12.162$\pm$0.084   \\
                           & walk         & 23.772$\pm$0.668 & 24.152$\pm$1.002 & 15.275$\pm$0.910 & 15.24$\pm$1.264    \\ \hline
\multirow{3}{*}{Walker}    & run          & 8.563$\pm$0.612  & 8.567$\pm$0.196  & 8.955$\pm$0.352  & 5.338$\pm$0.392    \\
                           & walk         & 6.876$\pm$1.103  & 8.334$\pm$1.613  & 7.231$\pm$1.101  & 3.664$\pm$0.276    \\
                           & stand        & 3.93$\pm$0.796   & 3.775$\pm$0.648  & 2.539$\pm$0.394  & 2.009$\pm$0.414    \\ \hline
\multirow{2}{*}{Cheetah}   & run          & 3.375$\pm$0.456  & 4.623$\pm$0.609  & 2.769$\pm$0.232  & 2.716$\pm$0.335    \\
                           & run backward & 2.737$\pm$0.229  & 4.07$\pm$0.164   & 2.244$\pm$0.118  & 1.981$\pm$0.002    \\ \hline
\end{tabular}
\caption{Distance to the fifth goal in multi-task multi-goal reaching.}
\label{tab:fifth}
\end{table}
\subsection{Finetuning Results on Model Scalability}
\begin{figure}[t!]
    \centering
    \includegraphics[width=\textwidth]{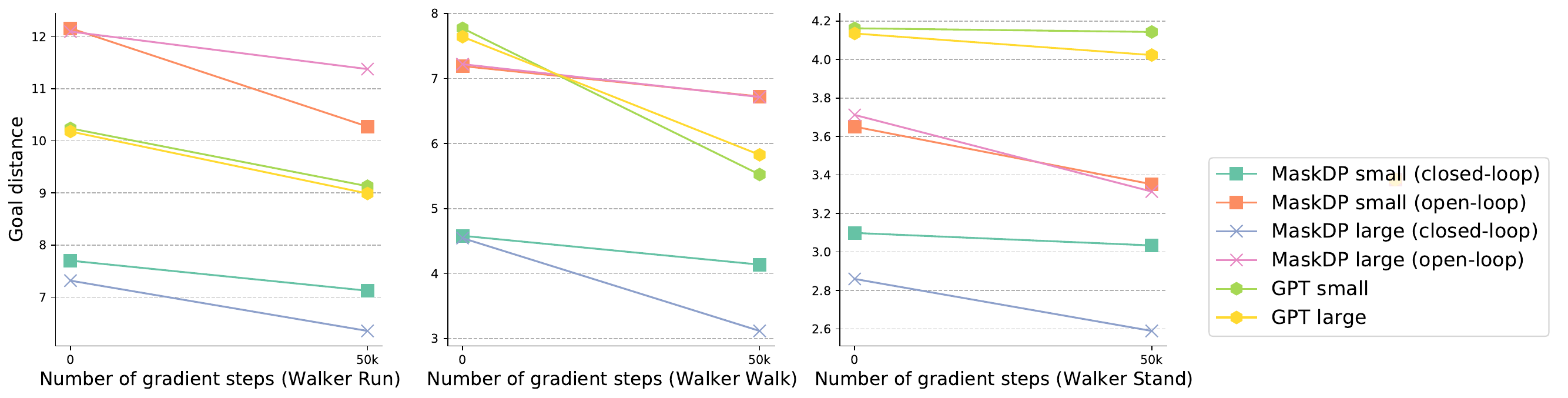}
    \caption{Model scalability on walker domain. X-axis represents number of gradient steps. With \ours{} pre-training, larger models outperform smaller models across all Walker tasks. In contrast, Goal-GPT does not have such properties.}
    \label{fig:walker_finetune}
\end{figure}

\begin{figure}[t!]
    \centering
    \includegraphics[width=\textwidth]{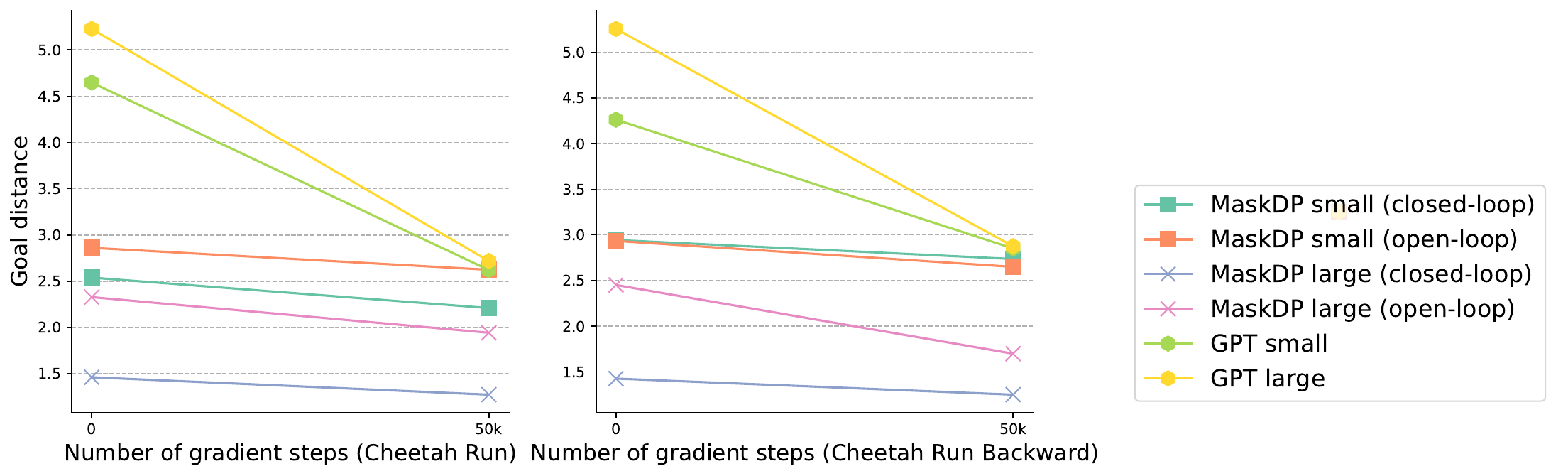}
    \caption{Model scalability on cheetah domain. X-axis represents number of gradient steps. With \ours{} pre-training, larger models outperform smaller models across all Walker tasks. In contrast, Goal-GPT does not have such properties.}
    \label{fig:cheetah_finetune}
\end{figure}

We pretrain \ours{} using the diverse multi-task mixed dataset, and finetune it using near-expert dataset on each task. In addition to the results in Figure~\ref{fig:scaling} on the Quadruped domain, we also provide results on the other two domains in Figure~\ref{fig:walker_finetune} and Figure~\ref{fig:cheetah_finetune}. Here ``small'' represents a model with 3 attention layers, while ``large'' represents 5 attention layers.

We can see the large model with closed-loop evaluation always performs the best, while for Goal-GPT the results are much worse, and the gain from the large model is not significant. 

